\documentclass[11pt,a4paper]{article}
\usepackage[hyperref]{emnlp2020}
\usepackage{times}
\usepackage{latexsym}
\usepackage{graphicx}
\usepackage{multirow}

\usepackage{xcolor}
\usepackage{amsmath}
\DeclareMathOperator*{\argmax}{arg\,max}

\usepackage{microtype}

\aclfinalcopy 
\def\aclpaperid{574} 

\title{SRLGRN: Semantic Role Labeling Graph Reasoning Network}

\author{Chen Zheng \\
  Michigan State University  \\
  \texttt{zhengc12@msu.edu} \\\And
  Parisa Kordjamshidi \\
  Michigan State University \\
  \texttt{kordjams@msu.edu} \\}

\date{}

\begin{document}

\maketitle
\begin{abstract}
This work deals with the challenge of learning and reasoning over multi-hop question answering (QA). We propose a graph reasoning network based on the semantic structure of the sentences to learn cross paragraph reasoning paths and find the supporting facts and the answer jointly. The proposed graph is a heterogeneous document-level graph that contains nodes of type sentence (question, title, and other sentences), and semantic role labeling sub-graphs per sentence that contain arguments as nodes and predicates as edges. Incorporating the argument types, the argument phrases, and the semantics of the edges originated from SRL predicates into the graph encoder helps in finding and also the explainability of the reasoning paths. Our proposed approach shows competitive performance on the HotpotQA distractor setting benchmark compared to the recent state-of-the-art models.

\end{abstract}

\section{Introduction}
Understanding and reasoning over natural language plays a significant role in artificial intelligence tasks such as Machine Reading Comprehension (MRC) and Question Answering (QA). Several QA tasks have been proposed in recent years to evaluate the language understanding capabilities of machines ~\cite{Rajpurkar2016SQuAD10,Joshi2017TriviaQAAL,Dunn2017SearchQAAN}. These tasks are single-hop QA tasks and consider answering a question given only one single paragraph.
Many existing neural models rely on learning context and type-matching heuristics~\cite{Weissenborn2017MakingNQ}. Those rarely build reasoning modules but achieve promising performance on single-hop QA tasks. The main reason is that these single-hop QA tasks are lacking a realistic evaluation of reasoning capabilities because they do not require complex reasoning.
\begin{figure}[ht!]
\centering
\includegraphics[width=0.42\textwidth,height=0.25\textheight]{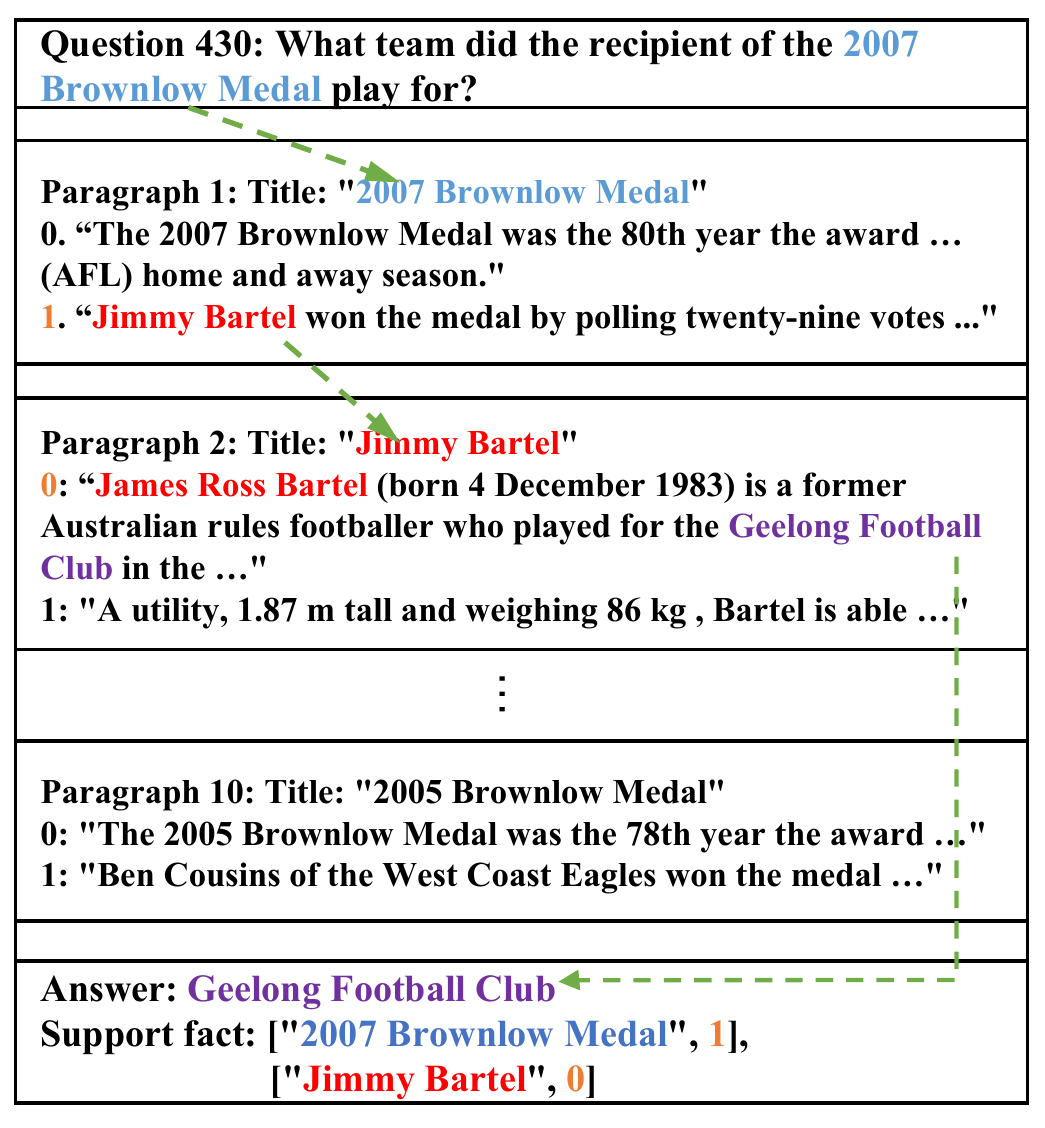}
\caption{An example of HotpotQA data.}
\label{fig:hotpotqa_example}
\end{figure}

Recently multi-hop QA tasks, such as HotpotQA ~\cite{Yang2018HotpotQAAD} and WikiHop~\cite{welbl2018constructing}, have been proposed to assess multi-hop reasoning ability.
HotpotQA task provides annotations to evaluate document level question answering and finding supporting facts. Providing supervision for supporting facts improves explainabilty of the predicted answer because they clarify the cross paragraph reasoning path.  
Due to the requirement of multi-hop reasoning over multiple documents with strong distraction, multi-hop QA tasks are challenging. 
Figure~\ref{fig:hotpotqa_example} shows an example of HotpotQA. Given a question and 10 paragraphs, only paragraph $1$ and paragraph $2$ are relevant. The second sentence in paragraph $1$ and the first sentence in paragraph $2$ are the supporting facts. The answer is ``Geelong Football Club''. 

Primary studies in HotpotQA task prefer to use a reading comprehension neural model~\cite{Min2019MultihopRC, Zhong2019CoarsegrainFC,Yang2018HotpotQAAD}. First, they use a neural retriever model to find the relevant paragraphs to the question. After that, a neural reader model is applied to the selected paragraphs for answer prediction. Although these approaches obtain promising results, the performance of evaluating multi-hop reasoning capability is unsatisfactory~\cite{Min2019MultihopRC}. 

To solve the multi-hop reasoning problem, some models tried to construct an entity graph using Spacy\footnote{\url{https://spacy.io}} or Stanford CoreNLP~\cite{Manning2014TheSC} and then applied a graph model to infer the entity path from question to the answer~\cite{Chen2019MultihopQA,Xiao2019DynamicallyFG,Clark2018SimpleAE,Fang2019HierarchicalGN}. However, these models ignore the importance of the semantic structure of the sentences and the edge information and entity types in the entity graph. To take the in-depth semantic roles and semantic edges between words into account here we use semantic role labeling (SRL) graph as the backbone of a graph convolutional network.
Semantic role labeling provides the semantic structure of the sentence in terms of argument-predicate relationships~\cite{He2018JointlyPP}. 
The argument-predicate relationship graph can significantly improve the multi-hop reasoning results. Our experiments show that SRL is effective in finding the cross paragraph reasoning path and answering the question.

Our proposed semantic role labeling graph reasoning network (SRLGRN) jointly learns to find cross paragraph reasoning paths and answers questions on multi-hop QA. In SRLGRN model, firstly, we train a paragraph selection module to retrieve gold documents and minimize distractor.
Second, we build a heterogeneous document-level graph that contains sentences as nodes (question, title and sentence), 
and SRL sub-graphs including semantic role labeling arguments as nodes and predicates as edges.
Third, we train a graph encoder to obtain the graph node representations that incorporate the argument types and the semantics of the predicate edges in the learned representations.
Finally, we jointly train a multi-hop supporting fact prediction module that finds the cross paragraph reasoning path, and answer prediction module that obtains the final answer. Notice that both supporting fact prediction and answer prediction are based on contextual semantics graph representations as well as token-level BERT pre-trained representations. The contributions of this work are as follows: 

\noindent{\bf 1)} We propose the SRLGRN framework that considers the semantic structure of the sentences in building a reasoning graph network. Not only the semantics roles of nodes but also the semantics of edges are exploited in the model.

\noindent{\bf 2)} We evaluate and analyse the reasoning capabilities of the semantic role labeling graph compared to usual entity graphs. 
The fine-grained semantics of SRL graph help in both finding the answer and the explainability of the reasoning path.

\noindent{\bf 3)} Our proposed model obtains competitive results on both HotpotQA (Distractor setting) and the SQuAD benchmarks.

\section{Related Work}
\begin{figure*}[!ht]
\centering
\includegraphics[width=0.95\textwidth,height=4cm]{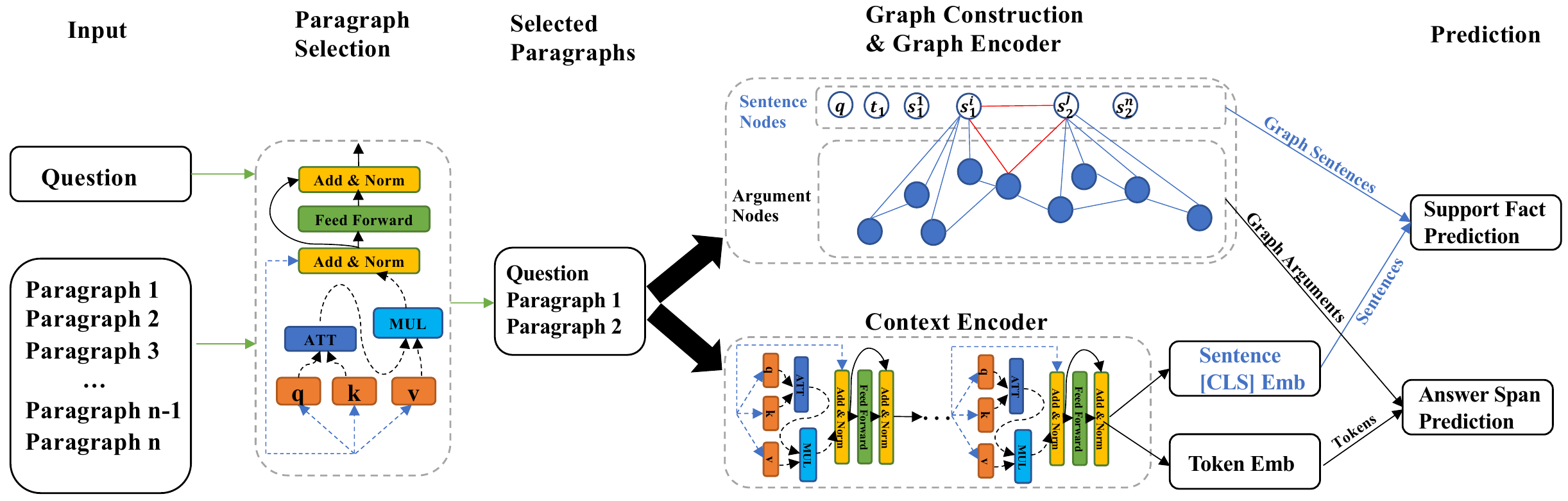}
\caption{Our proposed SRLGRN model is composed of Paragraph Selection, Graph Construction,  Graph Encoder, Supporting Fact prediction, and Answer Span prediction.}
\label{fig:architecture}
\end{figure*}

\subsection{Graph Models for Multi-Hop Reasoning}
Previous QA datasets, such as TriviaQA ~\cite{Joshi2017TriviaQAAL} and SearchQA ~\cite{Dunn2017SearchQAAN}, and MRC datasets, like SQuAD ~\cite{Rajpurkar2016SQuAD10}, rarely require sophisticated reasoning (such as cross paragraph reasoning) to answer the question and fail to provide ground-truth explanations for answers. 
Recently, WikiHop~\cite{welbl2018constructing} and HotpotQA~\cite{Yang2018HotpotQAAD} are two published multi-hop QA datasets that provide multiple paragraphs. Those QA datasets require a multi-hop reasoning model to learn the cross paragraph reasoning paths and predict the correct answer.

Most of the existing multi-hop QA models ~\cite{Tu2019SelectAA, Xiao2019DynamicallyFG, Fang2019HierarchicalGN} utilize graph based neural networks, such as graph attention network ~\cite{Velickovic2018GraphAN}, graph recurrent network ~\cite{Song2018AGM}, and graph convolutional network ~\cite{kipf2017semi}. 
Moreover, multi-hop QA models use different ways to construct entity graphs. Coref-GRN ~\cite{Dhingra2018NeuralMF} utilize co-reference resolution to build the entity graph.  MHQA-GRN ~\cite{Song2018ExploringGP} is an updated version of Coref-GRN that adds sliding windows. Entity-GCN ~\cite{Cao2019QuestionAB} builds the graph using entities and different types of edges called match edges and complement edges. DFGN ~\cite{Xiao2019DynamicallyFG} and SAE ~\cite{Tu2019SelectAA} construct entity graph through named entity recognition (NER).

In contrast to the above mentioned models, our SRLGRN builds a heterogeneous graph that contains a document-level graph of various sentences and replaces the entity-based graphs with argument-predicate based sub-graphs using SRL. 


\subsection{Semantic Role Labeling}
The goal of semantic role labeling is to capture argument and predicate relationships given a sentence, such as “who did what to whom.”  Several deep SRL models achieve highly accurate results in finding argument spans ~\cite{Zhou2015EndtoendLO, Tan2018DeepSR, Marcheggiani2017ASA, He2017DeepSR}. However, those models are evaluated based on given gold predicates. Therefore, some deep models~\cite{He2018JointlyPP, Guan2019SemanticRL} are proposed to recognize all argument-predicate pairs.
Recently, \citeauthor{shi2019simple} proposed a BERT Model for SRL and Relation Extraction.

\section{Model Description}
\label{sec:model}

Our proposed SRLGRN approach is composed of Paragraph Selection, Graph Construction, Graph encoder, Supporting Fact prediction, and Answer Span prediction modules. Figure~\ref{fig:architecture} shows the proposed architecture. In this section, we introduce our approach in detail and then explain how to train it with an efficient algorithm. 

\subsection{Problem Formulation}
Formally, the problem is to predict supporting fact $y_{SF}$ and answer span $y_{ans}$ given input question $q$ and candidate paragraphs. 
Each paragraph content $\mathcal{C}=\{t,s_1, \dots, s_n\}$ includes title $t$ and several sentences $\{s_1, \dots, s_n\}$.

\subsection{Paragraph Selection}
\label{sec:ps}

\begin{figure*}[!ht]
\centering
\includegraphics[width=0.9\textwidth,height=4.2cm]{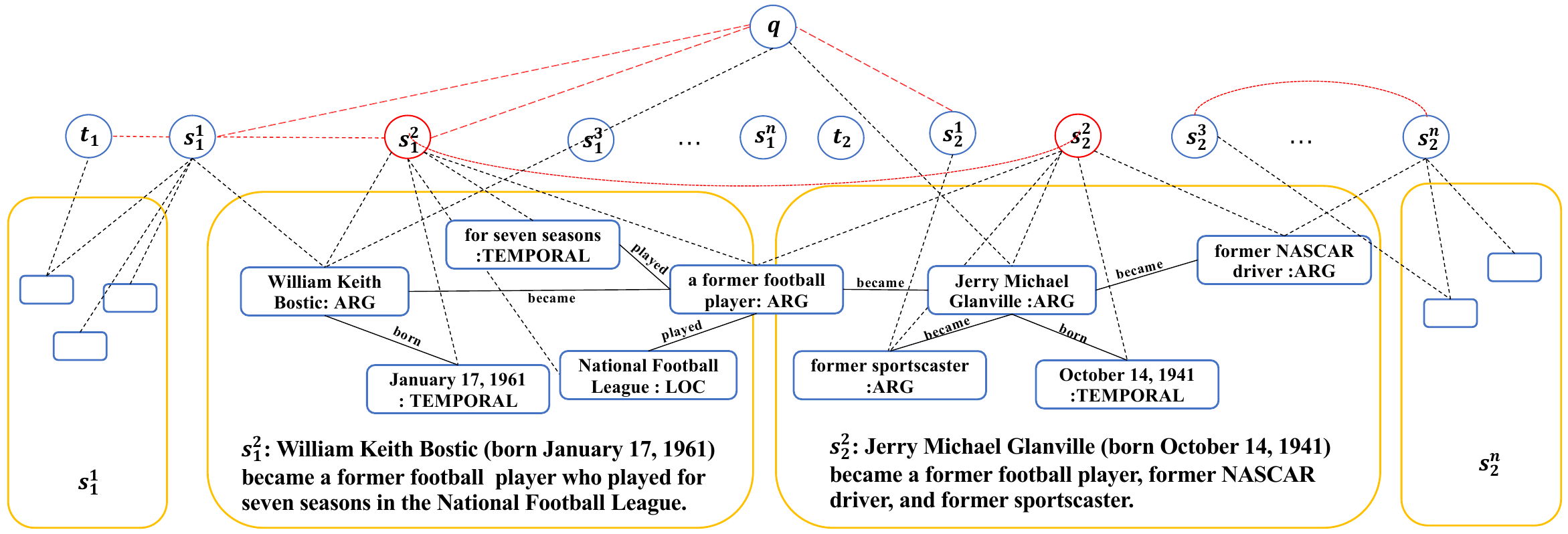}
\caption{An example of Heterogeneous SRL Graph. The question is ``Who is younger Keith Bostic or Jerry Glanville?'' The circles show the document-level nodes, i.e., sentences. The blue squares show the argument nodes. The argument nodes include argument phrase and argument type information. The solid black lines are semantic edges between two arguments carrying the predicate information. The black dashed lines show the edges between sentence nodes and  argument nodes. The red dashed lines show the edges between two sentences if there exists a shared argument (based on exact string match). The orange blocks are the SRL argument-predicate sub-graphs for sentences. $s_i^j$ means the $j$-th sentence from the $i$-th paragraph.}
\label{fig:srl}
\end{figure*}

Most of the paragraphs are distractors in the HotpotQA task~\cite{Yang2018HotpotQAAD}. 
SRLGRN can select gold documents and minimize distractors from given $N$ documents by a Paragraph Selection module.  
The Paragraph Selection is based on the pre-trained BERT model~\cite{devlin2018bert}. Our Paragraph Selection module has two rounds explained in section~\ref{sec:ps1} and section~\ref{sec:ps2}.

\subsubsection{First Round Paragraph Selection}
\label{sec:ps1}
For every candidate paragraph, we take the question $q$ and the paragraph content $\mathcal{C}$
as input: 
\begin{align}
Q_1 = [ [CLS];q;[SEP]; \mathcal{C} ],
\end{align}
where $Q_1$ represents the input, [CLS] and [SEP] are the same as BERT tokenizer process ~\cite{devlin2018bert}.
We feed input $Q_1$ to a pre-trained BERT encoder to obtain token representations. Then we use $BERT_{[CLS]}$ token representation as the summary representation of the paragraph.
Meanwhile, we utilize a two-layer MLP to output the relevance score, $y_{sel}$.
The paragraph which obtains the highest relevance score is selected as the first relevant context. We concatenate $q$ to the selected paragraph as $q_{new}$ for the next round of paragraph selection.

\subsubsection{Second Round Paragraph Selection}
\label{sec:ps2}
For the remaining $N-1$ candidate paragraphs, we use the same model as first round paragraph selection to generate a relevance score that takes $q_{new}$ and paragraph content as input. We call this process as second round paragraph selection.
Similar to section~\ref{sec:ps1}, one of the remaining candidate paragraphs with the highest score is selected. Afterwards, we concatenate the question and the two selected paragraphs to form a new context used as the input text for graph construction.

\subsection{Heterogeneous SRL Graph Construction}
\label{sec:graph_construction}
We build a heterogeneous graph that contains document-level sub-graph $\mathcal{S}$ and argument-predicate SRL sub-graph $Arg$ for each data instance. 
In the graph construction process, the document level sub-graph $\mathcal{S}$ includes question $q$, title $t_1$ and sentences $s_1^{1,\dots,n}$ from first round selected paragraph, and title $t_2$ and sentences $s_2^{1,\dots,n}$ from the second round selected paragraph, that is $\{q, t_1, s_1^1, \dots, s_1^n, t_2, s_2^1, \dots, s_2^n\} \in \mathcal{S}$. 
The argument-predicate SRL sub-graphs $Arg$, including arguments as nodes and the predicates as edges, are generated using AllenNLP-SRL model~\cite{shi2019simple}. Each argument node is the concatenation of argument phrase and argument type, including ``TEMPORAL'', ``LOC'', etc.

Figure~\ref{fig:srl} describes the construction of the heterogeneous graph. The heterogeneous graph's edges are added as follows: 
\noindent{\bf 1)} There will be an edge between a sentence and an argument if an argument appears in this sentence (the black dashed lines in Figure~\ref{fig:srl}); 
\noindent{\bf 2)} Two sentences $s_i$ and $s_j$ will have an edge if they share an argument by exact matching (the red dashed lines);
\noindent{\bf 3)} Two argument nodes  ${Arg}_i$ and ${Arg}_j$ will have an edge if a predicate exists between ${Arg}_i$ and ${Arg}_j$ (the black solid lines);
\noindent{\bf 4)} There will be an edge between the question and sentence if they share an argument (the red dashed lines).

Figure~\ref{fig:srl} shows an example of a heterogeneous SRL graph. $s_1^2$ and $s_2^2$ are connected because of a shared argument node ``a former football player: ARG''. Besides, the shared argument node has several semantic edges, such as ``played'' and ``became''. In this way, the shared argument node and other connected argument nodes have argument-predicate relationships. 

We create two matrices based on the constructed graph that we will use in section~\ref{sec:gcn}.
We build a predicate-based semantic edge matrix $K$ and a heterogeneous edge weight matrix $A$. The semantic edge matrix $K$ is a matrix that stores the word index of the predicates. 
We initialize all the elements of K with empty, $\emptyset$.
If two argument nodes ${Arg}_i$ and ${Arg}_j$ related to the same predicate, we add that predicate word index to $K_{({Arg}_i,{Arg}_j)}$. Sometimes, ${Arg}_i$ and ${Arg}_j$ are related to more than one predicate.

In the meantime, the heterogeneous edge weight matrix $A$ is a matrix that stores different types of edge weights. We divide the edges into three types: sentence-argument edges, argument-argument edges, and sentence-sentence edges. 

The weight of a sentence-sentence edge is $1$ when two sentences share an argument. Meanwhile, the weight of a sentence-argument edge is $1$ if there exists an edge between a sentence and an argument.
If two argument nodes have an edge, the weight can be calculated by point-wise mutual information (PMI) ~\cite{bouma2009normalized}. The reason we use PMI is that it can better explain associations between nodes compared to the traditional co-occurrence count method ~\cite{Yao2019GraphCN}. 

\subsection{Graph Encoder}
\label{sec:gcn}
Section~\ref{sec:graph_construction} introduces the detailed process of building a heterogeneous graph.
Next, we introduce the Graph Convolution Network ~\cite{kipf2017semi} to obtain the graph embeddings.
Graph Convolution Network (GCN) is a multi-layer network that
uses the graph input directly and generates embedding vectors of the graph.

Besides, GCN plays an essential role in incorporating higher-order neighborhood nodes and helps in capturing the structural graph information.
The SRL graph uses the semantic structure of the sentence to form the graph nodes and semantic edges, making the GCN's representation more explainable.
For instance, the GCN node vectors of document level sub-graph help in finding the supporting fact path, while GCN node vectors of argument-predicate level sub-graph help in identifying the text span of the potential answers. 
In this work, we consider a two-layer GCN to allow message passing operations and learn the graph embeddings. The graph embeddings are computed as follows:
\begin{align}
E_1 &= (D^{-\frac{1}{2}} A D^{-\frac{1}{2}} )  [ X_{Arg}; X_{\mathcal{S}} ]  W_1   ,      \\    
G &= (D^{-\frac{1}{2}}  A D^{-\frac{1}{2}} )  f(E_1)  W_2  ,       
\end{align}
where $E_1$ and $G$ are graph embedding outputs of two GCN layers that incorporate higher-order neighborhood nodes by stacking GCN layers. 
$f(x)$ is an activation function, $D$ is the degree matrix of the graph~\cite{kipf2017semi}, $A$ is the heterogeneous edge weight matrix, and $W_1$ and  $W_2$ are the learned parameters. $X$ represents node embeddings, including argument-predicate embedding $X_{Arg}$ and sentence embedding $X_{\mathcal{S}}$. Notice that each argument embedding $X_{Arg}^i$ is the concatenation of the argument node $Arg^i$ embedding and the average embedding of $K_{Arg}^i$.
Given $G$, we use $G_{\mathcal{S}}$ to represent document level node embeddings, and $G_{Arg}$ to represent argument-predicate level node embeddings. 

\subsection{Supporting Fact Prediction}

\begin{figure}[ht!]
\centering
\includegraphics[width=0.47\textwidth,height=2.7cm]{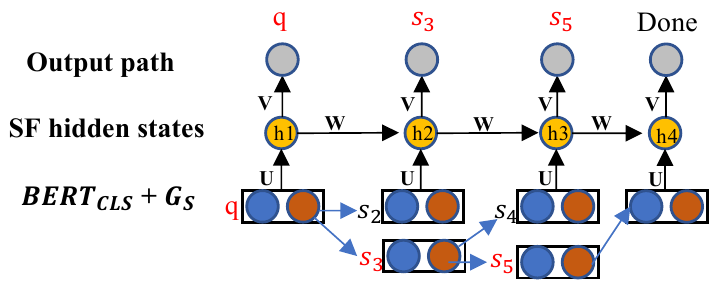}
\caption{An example of Supporting Fact Prediction.}
\label{fig:sup_fact}
\end{figure}

The goal of supporting fact (SF) prediction is to find the SF that is necessary to arrive at the answer.
Inspired by~\citeauthor{asai2020learning}, we utilize RNN with a beam search to find the best document-level SF path. This approach turns out to be effective for selecting the SF reasoning path. Notice that, our supporting fact prediction is not only based on BERT and RNN, but also incorporates document level graph node embeddings $G_{\mathcal{S}}$.

Formally, 
we use the concatenation of the graph sentence embedding, $G_\mathcal{S}$ (blue circles in Figure~\ref{fig:sup_fact}), and BERT's [CLS] token representation (orange circles) to represent the candidate sentence $X_{\mathcal{S}}^{cand}$:
\begin{align}
X_{\mathcal{S}}^{cand} &= [G_{\mathcal{S}}^{cand};BERT_{[CLS]}(q, \mathcal{S}_{cand})],
\end{align}
where $\mathcal{S}_{cand}$ represents the neighbors of the candidate sentence.
Afterwards, two feed-forward fully connected layers with activation functions determine whether $s_{cand}$ is an actual SF.
The process of selecting an SF is shown as follows:
\begin{align}
h_t &= \sigma(W h_{t-1}+U X_{\mathcal{S}}^{cand}+b_h), \\
o_t &= V h_t + b_o,
\end{align}
where $h_t$ is the hidden state of the RNN at the $t$-th SF reasoning step, $\sigma$ is the activation function. $W$, $U$, $V$, $b_h$ and $b_o$ are the parameters.

Finally, we use the beam search to output SF paths, choosing the highest scored path as our final supporting fact answer $y_\mathrm{SF}$:
\begin{align}
y_\mathrm{SF} &= \argmax_{ 1 \leq t\leq T } \prod o_t,
\end{align}
where $T$ is the maximum number of reasoning hops.
We penalize with the cross-entropy loss. More details are described in section~\ref{sec:objective}.

Figure~\ref{fig:sup_fact} shows an example of the predicted SF process. Based on the constructed heterogeneous graph, two sentence nodes have an edge if they share an argument. We start from question node $q$ as the first input sentence. Since $q$ is a unique input, 
we select $q$ as the first SF candidate.
In the second step, two candidate sentence nodes, $s_2$ and $s_3$ that are neighbor nodes of $q$, are chosen as the input. We separately feed $s_2$ and $s_3$ to the RNN layers. 
The sentence $s_3$ that obtains a larger logit score is selected as the second SF candidate of the reasoning path.
In the third step, $s_4$ and $s_5$ are neighbor nodes of the second SF, $s_3$. Then the model chooses $s_5$ as the third SF. In the end, $s_1$, $s_3$, and $s_5$ are the supporting facts. 

\subsection{Answer Span Prediction}
The goal of the answer span prediction module is to output ``yes'', ``no'', or answer span for the final answer. 
We firstly design an answer type classification based on BERT and an additional two fully connected feed-forward layers. 
If the highest probability of type classification is ``yes'' or ``no'', we directly output the answer. The input of type classification is
$BERT_{[CLS]}$. The answer type $y_{type}$ can be calculated as 
\begin{align*}
y_\mathrm{type} = MLP_\mathrm{type}([BERT_{[CLS]}]).
\end{align*}

If the answer is not ``yes'' or ``no'', we compute the logit of every token to find the start position $i$ and end position $j$ for answer span. The logit is calculated using BERT as the input given to two fully connected layers. 
The input token representation is the concatenation of BERT token representation $BERT_{tok}$ and graph embedding $G_{Arg}$.
The answer span $y_{ans}$ can be computed as
\begin{align}
y_\mathrm{ans} &= \argmax_{i,j,~i\leq j}y_{start}^i y_{end}^j, \\
y_\mathrm{start}^i  &= MLP_\mathrm{start}([BERT_{tok}^i;G_{Arg}^{i}]),\\
y_\mathrm{end}^i &= MLP_\mathrm{end}([BERT_{tok}^i;G_{Arg}^{i}]),
\end{align}
where $y_\mathrm{ans}$ is the index pair of (start position, end position), $y_\mathrm{start}^i$ represents the logit score of the $i$-th word as the start position, and $y_\mathrm{end}^i$ represents the logit score of the $i$-th word as the end position.

\subsection{Objective Function}
\label{sec:objective}
Inspired by~\citeauthor{Xiao2019DynamicallyFG} and~\citeauthor{Tu2019SelectAA}, the joint objective function includes the sum of cross-entropy losses for the span prediction $L_\mathrm{ans}$, answer type classification $L_\mathrm{type}$, and supporting fact prediction $L_\mathrm{SF}$. The loss function is computed as follows:
\begin{align*}
    L_\mathrm{joint} &= L_\mathrm{ans} + L_\mathrm{SF} + L_\mathrm{type} \\
                  &= \lambda_1(- y_\mathrm{start}\log{y_\mathrm{start}} - y_\mathrm{end}\log{y_\mathrm{end}}) \\
                  &- \lambda_2 y_\mathrm{SF} \log{y_\mathrm{SF}}- \lambda_3 y_\mathrm{type}\log{y_\mathrm{type}},
\end{align*}
where $\lambda_1$, $\lambda_2$, and $\lambda_3$ are weighting factors.

\section{Experiments and Results}

\begin{table*}[h!]
    \small
    \begin{center}
        \begin{tabular}{|c|c|c|c|c|c|c|} 
            \hline
            \multirow{2}{*}{\textbf{Model}} & \multicolumn{2}{c|}{\textbf{Ans}$(\%)$} & \multicolumn{2}{c|}{\textbf{Sup}$(\%)$} & \multicolumn{2}{c|}{\textbf{Joint}$(\%)$} \\
            & EM & F1 & EM & F1 & EM & F1 \\
            \hline
            Baseline Model \cite{Yang2018HotpotQAAD} & 45.60 & 59.02 & 20.32 & 64.49 & 10.83 & 40.16 \\
            KGNN \cite{Ye2019MultiParagraphRW} & 50.81 & 65.75 & 38.74 & 76.79 & 22.40 & 52.82 \\
            QFE \cite{Nishida2019AnsweringWS} & 53.86 & 68.06 & 57.75 & 84.49 & 34.63 & 59.61 \\
            DecompRC \cite{Min2019MultihopRC} & 55.20 & 69.63  & - & - & - & - \\
            DFGN \cite{Xiao2019DynamicallyFG} & 56.31 & 69.69 & 51.50 & 81.62 & 33.62& 59.82 \\
            TAP & 58.63 & 71.48 & 46.84 & 82.98 & 32.03 & 61.90 \\
            SAE-base \cite{Tu2019SelectAA} & 60.36 & 73.58 & 56.93 & 84.63 & 38.81 & 64.96 \\
            ChainEx \cite{Chen2019MultihopQA} & 61.20 & 74.11 & - & - & - & - \\
            HGN-base \cite{Fang2019HierarchicalGN} & - & 74.76 & - & \textbf{86.61} & - & \textbf{66.90} \\
            \hline
            SRLGRN-base & \textbf{62.65} & \textbf{76.14} & \textbf{57.30} & 85.83 & \textbf{39.41} & 66.37 \\
            \hline
        \end{tabular}
    \end{center}
    \caption{HotpotQA Result on Distractor setting. Except Baseline model, all models deploy BERT-base uncased as the pre-training language model to compare the performance. }
    \label{tab:hotpot_res}
\end{table*}

\subsection{Dataset}

We use the HotpotQA dataset ~\cite{Yang2018HotpotQAAD}, a popular benchmark for multi-hop QA task, for the main evaluation of the SRLGRN. 
Specifically, two sub-tasks are included in this dataset: Answer prediction and Supporting facts prediction. For each sub-task, exact match (EM) and partial match (F1) are two official evaluations that follow the work of ~\citeauthor{Rajpurkar2016SQuAD10}. 
A joint EM and F1 score are used to measure the final performance of both answer and supporting fact prediction. 
We evaluate the model on the Distractor Setting.
For each question in the Distractor Setting, two gold paragraphs and 8 distractor paragraphs, which are collected by a high-quality TF-IDF retriever from Wikipedia, are provided. Only gold paragraphs include ground-truth answers and supporting facts.
In addition, we use MRC datasets, Stanford Question-Answering Dataset (SQuAD) v1.1~\cite{Rajpurkar2016SQuAD10} and v2.0~\cite{Rajpurkar2018KnowWY}, to demonstrate the language understanding ability of our model.

\subsection{Implementation Details}
We implemented SRLGRN using PyTorch
We use a pre-trained BERT-base language model with 12 layers, 768-dimensional hidden size, 12 self-attention heads, and around $110M$ parameters ~\cite{devlin2018bert}.
We keep 256 words as the maximum number of words for each paragraph. For the graph construction module, we utilize a semantic role labeling model~\cite{shi2019simple} from AllenNLP\footnote{\url{https://demo.allennlp.org/semantic-role-labeling.}} to extract the predicate-argument structure. For the graph encoder module, we use 300-dimensional GloVe~\cite{Pennington2014GloveGV} pre-trained word embedding. The model is optimized using Adam optimizer~\cite{Kingma2015AdamAM}.

\subsection{Baselines}
\paragraph{Baseline Model}~\cite{Yang2018HotpotQAAD} makes use of~\citeauthor{Clark2018SimpleAE} approach. The model includes some neural modules that are based on self-attention and bi-attention~\cite{Seo2017BidirectionalAF}.

\paragraph{DFGN}~\cite{Xiao2019DynamicallyFG} is a strong baseline method for the HotpotQA task. DFGN builds an entity graph from the text. Moreover, DFGN includes a dynamic fusion layer that helps in finding relevant supporting facts.

\paragraph{SAE}~\cite{Tu2019SelectAA} is an effective Select, Answer and Explain system for multi-hop QA. SAE is a pipeline system that first selects the relevant paragraph and uses the selected paragraph to predict the answer and the supporting fact.

\subsection{Results}
Table~\ref{tab:hotpot_res} shows the results of HotpotQA (Distractor setting). We can observe the SRLGRN model exceeds most published results. Our model obtains a Joint Exact Matching (EM) score of 39.41$\%$ and Partial Matching (F1) score of 66.37$\%$ on joint performance. Our SRLGRN model has a significant improvement, about $28.58\%$ on Joint EM and $26.21\%$ on F1, over the  Baseline Model~\cite{Yang2018HotpotQAAD}.
Compared to the current published state of the art, SAE model~\cite{Tu2019SelectAA}, our model improves EM about $2.29\%$ and F1 about $2.56\%$  on Answer performance and $1.41\%$ of F1 on Joint performance.
We can observe that F1 of answer span prediction is better than the current SOTA. The reason is that our model not only uses token-level BERT representation, but also uses graph-level SRL node representations.

Our framework provides an effective way for multi-hop reasoning taking the advantages of the SRL graph model and powerful pre-trained language models. In the following section, we give a detailed analysis of the SRLGRN model.

\section{Analysis}
\paragraph{Effect of SRL Graph.}
The SRL graph extracts argument-predicate relationships, including in-depth semantic roles and semantic edges. The constructed graph is the basis of reasoning as the inputs of each hop are directly selected from the SRL graph, as shown in Figure~\ref{fig:sup_fact}. The SRL graph significantly improves the completeness of the graph network, that is, providing sufficient semantic edges to cover reasoning paths, see Figure~\ref{fig:srl}. 

Compared to the NER graph in the previous models ~\cite{Xiao2019DynamicallyFG}, the proposed SRL graph covers the $86.5\%$ of complete reasoning paths for the data samples. The NER graph of DFGN is incomplete and can only cover $68.7\%$ of the reasoning paths~\cite{Xiao2019DynamicallyFG}.
The graph completeness is one major reason that the SRLGRN model has higher accuracy than other published models. As shown in Table~\ref{tab:hotpot_res}, the SRLGRN improves $5.79\%$ on joint EM and $6.55\%$ on joint F1 over DFGN, which is based on the NER graph.
\begin{table}[ht!]
    \begin{center}
    \small
        \begin{tabular}{|c|c|c|c|}
            \hline
            \multirow{2}{*}{\textbf{Ablation}} & \multirow{2}{*}{\textbf{Model}} & \multicolumn{2}{c|}{\textbf{Ans}$(\%)$} \\
            & & EM & F1 \\
            \hline
            
            \multirow{3}{*}{\textbf{Graph}}
            & w/o graph & 53.06 & 67.68 \\ 
            \cline{2-4}
            & \multirow{1}{*}{w/o Argument type} & &  \\
            &  and Semantic edge & 60.10 & 73.24 \\
            \hline
            \textbf{Joint} & w/o joint training & 58.50 & 71.58 \\
            \hline
            
            \multirow{2}{*}{\textbf{Language}} & ALBERT-base & 59.87  & 74.20  \\
            &  BERT-base  & 62.65 & 76.14 \\
            \hline
        \end{tabular}
    \end{center}
    \caption{SRLGRN ablation study on HotpotQA.}
    \label{tab:hotpot_graph_ana}
\end{table}

To evaluate the effectiveness of the types of semantic roles and the edge types, we perform an ablation study. 
First, we removed the whole SRL graph. Second, we removed the predicate based edge information from the SRL graph. 
Table~\ref{tab:hotpot_graph_ana} shows the results. The complete SRLGRN improves $8.46\%$ on  F1 score compared to the model without the SRL graph. 
The model loses the connections used for multi-hop reasoning if we remove the SRL graph and only use BERT for answer prediction. 

We also observe that the F1 score of answer span prediction decreases $2.9\%$ if we did not incorporate semantic edge information and argument types. The reason is that removing predicate edges and argument types will destroy the argument-predicate relationships in the SRL graph and breaks the chain of reasoning.
For example, in Figure~\ref{fig:srl}, the main arguments of the two supporting facts in $s_1^2$ and $s_2^2$ (William and Jerry) are connected with a predicate edge, ``born'', to the temporal information necessary for finding the answer. Both ``born'' edge and the adjunct temporal roles are the key information in the two sentences to find the final answer to this question. The shared ARG node, ``football player'', also helps to connect the line of reasoning between the two sentences.
These two results indicate that both semantic roles and semantic edges in the SRL graph are essential for the SRLGRN performance.

In a different experiment, we tested the influence of the joint training of the supporting facts and answer-prediction. As shown in Table~\ref{tab:hotpot_graph_ana}, the performance will decrease by $4.56\%$ when we did not train the model jointly.

\paragraph{Effect of Language Models.} 
\label{sec:LM_analysis}
We use two recent and widely-used pre-trained language representation models, BERT and ALBERT ~\cite{Lan2020ALBERTAL}. The last two lines of Table~\ref{tab:hotpot_graph_ana} show the  results.
Although BERT achieves relatively better performance, ALBERT architecture has significantly fewer parameters ($18$x) and is faster (about $1.7$x running time) than BERT. In other words, ALBERT reduces memory consumption by cross-layer parameter sharing, increases the speed, and obtains a satisfactory performance.

\paragraph{Effect of SRLGRN on Single-hop QA.}
We evaluate the SRLGRN (excluding the paragraph selection module) on SQuAD~\cite{Rajpurkar2016SQuAD10} to demonstrate its reading comprehension ability. 
We evaluate the performance on both SQuAD v$1.1$ and SQuAD v$2.0$.
Table~\ref{tab:squad1} describes the comparison results with several baseline methods on SQuAD v$1.1$. Our model obtains a $1.8\%$ improvement over BERT-large, and a $1.6\%$ improvement over BERT-large+TriviaQA~\cite{devlin2018bert}. 
\begin{table}[ht!]
    \begin{center}
    \small
        \begin{tabular}{|c|c|c|}
            \hline
            \multirow{2}{*}{\textbf{Model}} & \multicolumn{2}{c|}{\textbf{Ans}$(\%)$} \\
            & EM & F1 \\
            \hline
            Human & 82.3 & 91.2 \\
            \hline
            BERT-base  & 80.8 & 88.5  \\
            BERT-large & 84.1 & 90.9  \\
            BERT-large+TriviaQA & 84.2 & 91.1  \\
            \hline
            BERT-large+SRLGRN & 85.4 & 92.7 \\
            \hline
        \end{tabular}
    \end{center}
    \caption{SQuAD v1.1 performance.}
    \label{tab:squad1}
\end{table}

\noindent We further test the SRLGRN on SQuAD v$2.0$. The main difference 
is that SQuAD v$2.0$ combines answerable questions (like SQuAD v$1.1$) with unanswerable questions ~\cite{Rajpurkar2018KnowWY}. 
Table~\ref{tab:squad2} shows that our proposed approach improves the performance for SQuAD benchmark compared to several recent strong baselines.

\begin{table}[ht!]
    \small
    \begin{center}
        \begin{tabular}{|c|c|c|}
            \hline
            \multirow{2}{*}{\textbf{Model}} & \multicolumn{2}{c|}{\textbf{Ans}$(\%)$} \\
            & EM & F1 \\
            \hline
            Human & 86.3 & 89.0 \\
            \hline
            ELMo+DocQA~\cite{Rajpurkar2018KnowWY} & 65.1 & 67.6 \\
            BERT-large~\cite{devlin2018bert} & 78.7 & 81.9  \\
            SemBERT~\cite{Zhang2019SemanticsawareBF} & 84.8 & 87.6  \\
            \hline
            BERT-large+SRLGRN & 85.8 & 87.9 \\
            \hline
        \end{tabular}
    \end{center}
    \caption{SQuAD v2.0 performance.}
    \label{tab:squad2}
\end{table}

We recognize that our SRLGRN improves $7.1\%$ on EM compared to the robust BERT-large model and improves $1.0\%$ on EM compared to SemBERT~\cite{Zhang2019SemanticsawareBF}. 
The two experiments on SQuAD v$1.1$ and SQuAD v$2.0$ demonstrate the significance of SRL graph and the graph encoder.

\begin{table}[ht!]
    \small
    \begin{center}
    \resizebox{0.48\textwidth}{25mm}{
        \begin{tabular}{|c|c|c|} 
            \hline
            Error Type & Model Prediction & Label \\
            \hline
            \multirow{5}{*}{\textbf{Synonyms}}  & washington dc & district of columbia \\
                                                & sars  &  severe acute \\
                                                &  & respiratory syndrome\\
                                                & ey  &  ernst young \\
                                                & writer & author \\
            \hline
            \multirow{3}{*}{\textbf{MLV}}  & australian & australia \\
                                                   & hessian  &  hessians   \\
                                                  & mcdonald’s, co & mcdonalds \\
                                                
            \hline
            \multirow{3}{*}{\textbf{Month-Year}}  & 1946  &  1945 \\
                                                  &  25, november, 2015  &  3, december  \\
                                                  &  10, july, 1873  &  1, september, 1864  \\
                                                  
            \hline
            \multirow{3}{*}{\textbf{Number}}  & 11  &  10 \\
                                              & fourth &  4 \\
                                              & 2402   &  5922  \\
            \hline
            \textbf{External}  & Coker  &  NCAA I \\
            \textbf{Knowledge}    & & FBS football   \\
            \hline
            \multirow{3}{*}{\textbf{Other}}  & taylor, swift  &  usher   \\
                                             & film  &  documentary  \\
                                             & fourteenth  &  500th episode  \\
            
            \hline
        \end{tabular}
    }
    \end{center}
    \caption{Error types on HotpotQA dev set.}
    \label{tab:error_analysis}
\end{table}



\begin{figure*}[!ht]
\centering
\includegraphics[width=0.95\textwidth,height=5.8cm]{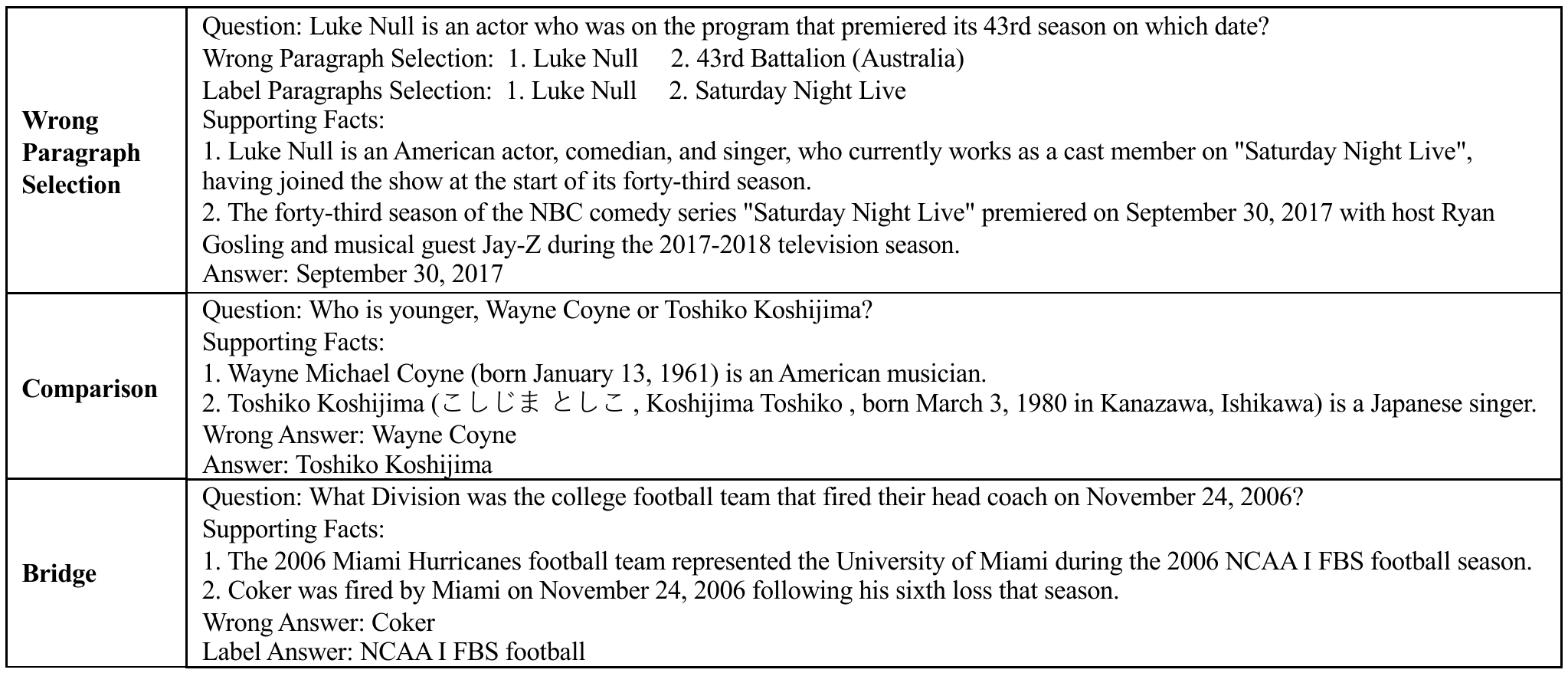}
\caption{ Failing cases on our proposed SRLGRN framework.  }
\label{fig:case_study}
\end{figure*}

\section{Error Analysis}
\paragraph{Synonyms} are the most frequent cause of the reported errors in many cases where the predicted answer is semantically correct.
As shown in the first row of the Table~\ref{tab:error_analysis}, 
our predicted answer and gold label have the same meaning.
For example, SRLGRN predicts "sars", while the label is "severe acute respiratory syndrome." We know that "sars" is the abbreviation of the gold label.

\paragraph{Minor Lexical Variation (MLV)}
is another major cause of mistakes in the SRLGRN model.  As shown in the second row of Table~\ref{tab:error_analysis}, our model's predicted answer is "australian", while the gold label is "australia". Many wrong predictions occur in the singular noun versus plural noun selection.


\paragraph{Paragraph Selection} 
is a small portion of errors in the SRLGRN model. As shown in Figure~\ref{fig:case_study}, the model chooses a wrong paragraph ``43rd Battalion''. The reason is that ``43rd Battalion'' is a distractor although ``43rd'' appears in the question. 
The paragraph ``Saturday Night Live'' is the correct relevant paragraph that includes ``forty-third season'' and the answer. To resolve this issue in the future, we will try to combine our model with an IR system designed for multi-hopQA similar to the Multi-step entity-centric model for multi-hop QA in~\cite{Godbole2019MultistepEI}.

\begin{figure}[!ht]
\centering
\includegraphics[width=0.45\textwidth,height=6.5cm]{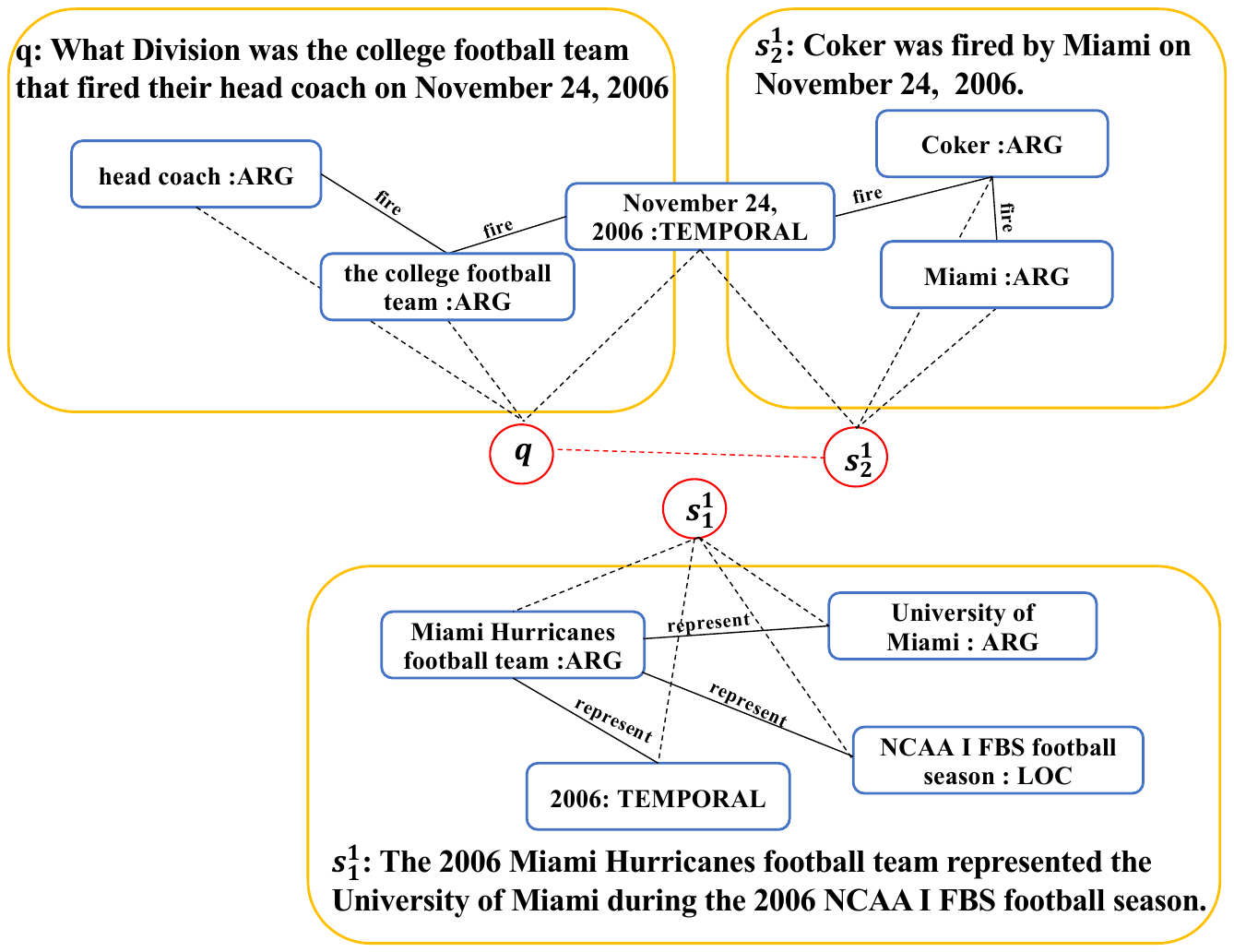}
\caption{ The ``Bridge'' failing case that SRL fails to lead to the correct answer. The meaning of different lines and node colors are the same as Figure~\ref{fig:srl}.}
\label{fig:srl_no_chain}
\end{figure}

\paragraph{Comparison and Bridge} are two types of reasoning that are needed for answering HotpotQA questions.
``Bridge” reasoning predicts the answer by connecting arguments to the line of reasoning that leads to the final answer. ``Comparison” reasoning predicts the answer (that is, yes, no, or a text span) by comparing two arguments.

SRLGRN sometimes obtains wrong predictions in the ``Comparison” reasoning when the questions are related to ``Month-year'' and ``Number''. Our qualitative error analysis showed that SRLGRN graph leads to a wrong answer when two or more argument nodes of a same type, such as ``TEMPORAL'' type, are connected to one node in the graph. Moreover, We notice that the SRLGRN sometimes makes inconsistent errors. For example, in the ``Comparison'' failing cases of Figure~\ref{fig:case_study}, we predict the wrong answer ``Wayne Coyne''. However, we received the correct answer after replacing the word ``younger'' with ``older''.

Moreover, the ``Bridge'' type needs external knowledge in the HotpotQA task. As is shown in ``Bridge'' failing cases of Figure~\ref{fig:case_study}, the selected paragraphs do not show the relation between ``Coker'' and ``Miami Hurricanes football team''. Figure~\ref{fig:srl_no_chain} describes the SRL construction based on this failing case. The second supporting fact and the question have the same temporal argument node ``November 24, 2006''. However, there is no chain between the first supporting fact and the second supporting fact due to the lack of the external knowledge  that can connect ``Coker'', ``coach'' and ``Miami Hurricanes football team''. Therefore, the isolated reasoning chain leads to a wrong answer.

\section{Conclusion}
We proposed a novel semantic role labeling graph reasoning network (SRLGRN) to deal with multi-hop QA. 
The backbone graph of our proposed graph convolutional network (GCN) is created based on the semantic structure of the sentences. In creating the  edges and nodes of the graph, we exploit a semantic role labeling sub-graph for each sentence and connect the candidate supporting facts. The cross paragraph argument-predicate structure of the sentences expressed in the graph provides an explicit representation of the reasoning path and helps in both finding and explaining the multiple hops of reasoning that lead to the final answer.
SRLGRN exceeds most of the SOTA results on the HotpotQA benchmark. Moreover, we evaluate the model (excluding the paragraph selection module) on other reading comprehension benchmarks. Our approach achieves competitive performance on SQuAD v$1.1$ and v$2.0$.

\section*{Acknowledgments}
This project is supported by National Science Foundation (NSF) CAREER award $\#$1845771 and (partially) supported by the Office of Naval Research grant $\#$N00014-19-1-2308.
We thank the anonymous reviewers for their thoughtful comments.  

\bibliographystyle{acl_natbib} 
\bibliography{emnlp2020}

\end{document}